\documentclass[letterpaper, 10 pt, conference]{ieeeconf}
\IEEEoverridecommandlockouts
\usepackage{cite}
\usepackage{amsmath,amssymb,amsfonts}
\usepackage{algorithmic}
\usepackage{graphicx}
\usepackage{textcomp}
\usepackage{xcolor}
\usepackage{comment}
\usepackage{subcaption}
\def\BibTeX{{\rm B\kern-.05em{\sc i\kern-.025em b}\kern-.08em
    T\kern-.1667em\lower.7ex\hbox{E}\kern-.125emX}}
\usepackage{lmodern}
\usepackage{siunitx}
\usepackage{booktabs}
\usepackage{etoolbox}
\usepackage[font=small]{caption}
\usepackage{graphicx}
\usepackage[export]{adjustbox}
\usepackage{multirow}
\usepackage{balance}
\newcommand{\cfg}{\mathbf{x}}
\newcommand{\Cfg}{\mathbf{X}}
\DeclareMathOperator\sign{sgn}
\makeatletter
\let\NAT@parse\undefined
\makeatother
\usepackage[bookmarks=false,pagebackref=true,breaklinks=true,colorlinks,linkcolor=blue,linktocpage=true,citecolor=blue,urlcolor=cyan,pagebackref=true,]{hyperref}
\begin{document}

\title{Data-driven Actuator Selection for Artificial Muscle-Powered Robots
}

\author{Taylor West Henderson$^{1}$, Yuheng Zhi$^{1}$, Angela Liu$^{1}$, and Michael C. Yip$^{1}$
\thanks{*This research is funded by NSF Award \#1830403.}
\thanks{$^{1}$Taylor West Henderson, Yuheng Zhi, Angela Liu and Michael C. Yip are with the Department of Electrical and Computer Engineering, University of California San Diego, La Jolla, CA 92093 USA
        {\tt\small \{tjwest, yzhi, anliu, yip\}@ucsd.edu}}%
} 

\maketitle

\begin{abstract}
Even though artificial muscles have gained popularity due to their compliant, flexible and compact properties, there currently does not exist an easy way of making informed decisions on the appropriate actuation strategy when designing a muscle-powered robot; thus limiting the transition of such technologies into broader applications.
What's more, when a new muscle actuation technology is developed, it is difficult to compare it against existing robot muscles. 
To accelerate the development of artificial muscle applications, we propose a data-driven approach for robot muscle actuator selection using \textit{Support Vector Machines} (SVM).
This first-of-its-kind method gives users insight into which actuators fit their specific needs and actuation performance criteria
, making it possible for researchers and engineers
with little to no prior knowledge of artificial muscles to 
focus on application design. It also provides a platform to benchmark existing, new or yet-to-be discovered artificial muscle technologies. We test our method on unseen existing robot muscle designs to prove its usability on real-world applications. We provide an open-access, web-searchable interface for easy access to our models that will additionally allow for the continuous contribution of new actuator data from groups around the world to enhance and expand these models.
\end{abstract}

\section{Introduction}
Artificial muscle actuators comprise a class of actuators that closely mimic the properties of biological muscles. As compliant, soft, and flexible robots have gained in popularity, so too have these biomimetic actuators \cite{Tondu_AM}, \cite{jun_review}. These robot muscle technologies span a wide variety of materials and configurations, including such notable examples as shape-memory alloys (SMAs), dielectric elastomers (DEAs), super-coiled polymers (SCPs), piezoelectric actuators (PZTs), and soft fluidic actuators (SFAs) (e.g. McKibben actuators) \cite{DEA, Yip2017, Softfluidic}. 
These actuators range from micro- to macroscopic scales and offer unique muscle-like properties including controlled compliances, large bandwidth ranges, high power-to-weight ratios, and compact muscle-like form factors \cite{AM_GENERAL_1, Haines2014, Yip2017}. When compared to traditional actuation methods such as motors or hydraulics, these artificial muscles and their desirable properties are vital in realizing more compliant, compact, and safe robotic designs and operations \cite{hexapod,pneumatic_application_GreerICRA2017,ankle-foot,crawling_robot}.

\begin{figure}
    \centering
    \includegraphics[width=.98\linewidth]{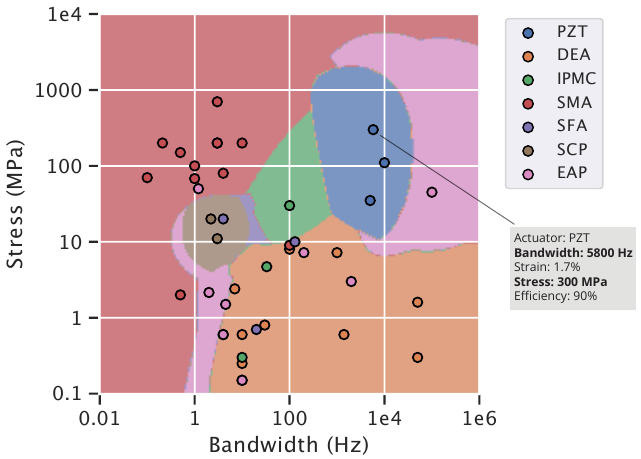}
    \caption{A large-scale data-driven machine learning strategy is proposed to help select artificial muscle actuators for user-specified robot performance constraints. The scattered points are sample actuators from a dataset collected via a comprehensive literature review, and the colors present the class boundaries. Actuator technologies shown across this new \textit{performance space} demonstrate key areas where they excel. The highlighted sample is from \cite{jani2014review}.} 
    \label{fig:opening}
\end{figure}

Despite decades of materials research focused on discovering new as well as improving existing muscle actuators, their transition into more widespread use in robot systems and applications has been relatively slow to date. Though they belong to a unifying class of \textit{artificial muscle actuators}, these biomimetic actuators have historically been considered dissimilar and distinct technologies as they vary widely in terms of material, configuration, and scale; this induces significant challenges for robot designers to overcome \cite{jun_review}. For one, when presented with the broad class of \textit{artificial muscle actuators}, it is unclear to the typical robot designer how to select or even compare muscle actuators for use in their specific application. Additionally, there are few resources available to academic researchers that allow them to compare the relative performance of their muscle actuation strategies with state-of-the-art developments.



In this paper, we propose a data-driven method of performing muscle selection over multi-dimensional performance criteria that incorporates empirical data derived from prior work (Fig. \ref{fig:opening}). 
We introduce the concept of a \textit{performance space} for artificial muscle actuators that includes relevant input and output parameters based on evidence from prior literature. 
This first-of-its-kind benchmarking strategy will not only characterize existing robot muscle technologies, but will allow new robot muscles to demonstrate their equivalency or novelty compared to the breadth of available actuators in an accelerated way, enabling high-performance robot muscles to emerge naturally. Fundamentally, this data-driven approach is key to lowering the barriers to entry in developing muscle-powered machines for current and future researchers, and hobbyists alike. We further test our proposed model on real-world robotics applications to ascertain its efficacy in selecting relevant muscle actuator types for a given set of performance specifications.  Finally, we provide an open-access web toolkit that utilizes our selection model and will proliferate and democratize the characterization of these robot muscles to a wider audience. 


\section{Background}
Traditional forms of actuation, such as electric motors and hydraulic actuators, are widely utilized by the robotics community to produce movements for robots and machines. While these actuators can achieve accurate positioning and produce large forces, they differ from natural muscles by often exhibiting inertia and friction while requiring substantial mass and large volumes \cite{mobile_robots, friction, hydraulics}. On the other hand, artificial muscle actuators provide an alternative that can generate muscle-like movements in a compliant, lightweight, and small form factor \cite{Tondu_AM, reviewSMA}. These robot muscles have shown immense promise as viable driving mechanisms for biomimetic robotic applications including robot manipulators, prostheses, exoskeletons, medical robots, and soft robots \cite{pneumatic_application_GreerICRA2017,ankle-foot,crawling_robot,orthoses}.


One obstacle related to this technology is the uncertainty regarding which properties of robot muscles should be considered as standard metrics of importance. Robot muscles have a large parameter space that is determined by factors such as the actuator configuration, fabrication procedures, actuation mechanism, and the external environment. These include compliance, bandwidth, precision, strain, robustness, cost, as well as others. It is therefore difficult to define a small set of metrics that are representative of the overall performance of robot muscles. For example, SCP actuators can generate a large untwisting torque under stimuli \cite{Haines2016}. A twisting motion is in general difficult to obtain using other robot muscles, so it would not qualify as a common performance metric for robot muscles. Bandwidth, on the other hand, is a common parameter across all types of muscle actuators regardless of the scale and motion and can be considered a standard metric for robot muscles \cite{reviewSMA,pico}.

In addition, the overall correlation of the properties must be understood, as many muscle actuator properties are coupled together. For example, achieving a larger bandwidth often results in the reduction of precision and strain \cite{variable_stiffness,meshworm,Haines2016}. Due to these complexities, significant effort is required to analyze and utilize different muscle actuators. In particular, when new actuation technologies are discovered, such as fluid-driven origami-inspired artificial muscles \cite{origami}, soft, robust composite material \cite{composite}, and Peano-HASEL (hydraulically amplified self-healing electrostatic) actuators \cite{peano}, robotic researchers and engineers are often required to understand and effectively compare nuanced robot muscle technologies that may be outside of their scope of knowledge. With this overhead to employ robot muscles, there have been few muscle-powered robots that are actually developed with complex motion capabilities.

\section{Methods}
The following section covers two major components of our methodology: (A) how we collect unstructured data for robot muscle actuators in extensive publications and studies and normalize it in order to apply machine learning algorithms; (B) how we define a data-driven muscle \textit{performance space} using SVMs to benchmark robot muscle actuators and provide guidance in actuator selection.

\subsection{Data Collection and Normalization}\label{sec:data}
One of the major hurdles to overcome in creating a unifying data-driven selection strategy for robot muscles is the acquisition of relevant data. 
As no centralized database of multiple muscle actuator parameters currently exists, we have created one. The set of quantitatively defined features for robot muscle usage must be determined by a comprehensive literature survey, spanning a valid set of features for robot muscles with different mechanisms, scales, and types of motion being generated. We have sampled the values of the muscle features from publications and studies listed in our review paper \cite{jun_review}, as well as other recent publications\footnote{{\href{http://robotmuscletoolkit.herokuapp.com/references}{http://robotmuscletoolkit.herokuapp.com/references}}}. While much of the data collected for each paper comes directly from tables and from within the body of the text, we also utilized WebPlotDigitizer \cite{webplotdigit} to 
collect viable data from plots when publications included plots of useful parameters instead of data points or tables. 

This comprehensive literature survey helped us identify five major features that define the performance of artificial muscles: strain, stress, bandwidth, efficiency, and power density. Later we use these universal actuator parameters to guide the definition of our \textit{performance space}. In this paper, the whole dataset of muscle parameters is denoted as $\Cfg \in \mathbb{R}^{N\times M}$, and the $i$-th sample is $\cfg_i \in \mathbb{R}^{M}$. $M=5$ is the maximum number of \textit{features} collected in this dataset. A positive integer number is assigned to each considered \textit{class} of actuators: PZT, DEA, IPMC (ionic polymer metal composites), SMA, SFA, SCP, and EAP (electroactive polymers other than DEAs and IPMCs). 
All considered classes form a set $\mathcal{C}=\{1, ..., C\}$, where $C=7$. The class labels corresponding to the actuators in the dataset are collected in $\mathbf{y} \in \mathcal{C}^{N}$, whose $i$-th instance is denoted as $y_i \in \mathcal{C}$. These types of actuators are considered since they have the most data readily available. However, it is worth noting that the proposed modeling method is designed to scale to more types of actuators as well as more types of performance metrics provided sufficient data, without loss of generality.


To effectively compare our actuator features both visually and through machine learning algorithms, we apply two normalization strategies to them. The first is to take the natural logarithm of each feature,
\begin{align}
    \Cfg_{\ln, ij} := \ln \Cfg_{ij},\quad \forall i\in[1,N], j\in[1,M].
\end{align}
Due to the differing mechanisms of various muscle actuators, the values for certain features may vary by orders of magnitude across actuator types. Taking the logarithm can distribute the data more uniformly and make it easier to handle for machine learning models. 
We then shift all features such that their means become zero, and scale them to have unit variances,
\begin{align}
    \Cfg'_{ij} := \frac{\Cfg_{\ln,ij} - \mu_j}{\sigma_j},\quad \forall i\in[1,N], j\in[1,M],
\end{align}
where $\mu_j,~\sigma_j$ stand for the mean and standard deviation of the $j$-th feature in $\Cfg_{\ln}$.

\subsection{Defining a Muscle Performance Space}
To accurately model the full class of artificial muscle actuators, we utilize a data-driven approach to quantitatively define features that are valid for robot muscles with different mechanisms, scales, and types of motion generation.

Using our chosen features, the robot muscles will be distinguished from one another through the accessible kernel support vector machine (SVM) method \cite{SVM}. The SVM method is chosen due to the sparsity of our compiled data. 
Since we anticipate the actuator feature data will not be linearly separable, we employ the radial basis function (RBF) as the kernel function $k(\cdot, \cdot)$ for our SVM algorithm.

One major challenge comes from partially missing data. Despite the extensive effort to collect data and carefully select major features, there is still a considerable number of missing features in some samples, which is reported in Sec.~\ref{sec:res_data}. Since the completion of missing actuator features remains an open problem and may be a direction of future work, instead of training SVMs on all 5 features at once, we train each SVM with only 2 input variables on samples with both valid features. In other words, an SVM is trained using each possible pair of features out of the 5 major features selected, producing $L = C_5^2=10$ models in total. Each model has $N_l$ valid training samples, $l\in \{1,...,L\}$. 
This technical choice is the consequence of a trade-off between the number of input features and valid training samples. It also provides an additional benefit of making visualization of the performance space straight-forward and intuitive. 
On the contrary, if the SVMs were trained using all 5 features, there would exist significantly less number of valid samples that have all 5 features available; also, it would be difficult to visualize a 5D performance space.

Even if we use only 2 features at a time, we note that there are 7 classes of actuators to be classified while the output of the standard SVM is binary in nature. Thus, we reduce our single multi-class problem into multiple binary classification problems. We build these binary support vector classifiers to distinguish between every pair of classes using the one-vs-one policy (a.k.a. \textit{ovo}), producing $K = C_7^2=21$ vanilla sub-SVMs for each pair of features. Each trained sub-SVM$_k$ contains a set of weights associated to its training samples $\mathbf{w}_{lk} = [w_{lk}^1, w_{lk}^2, ..., w_{lk}^{N_l}] \in \mathbb{R}^{N_l}$ and a bias term $b_{lk}\in \mathbb{R}$. $l$ here indicates it uses the $l$-th pair of features. Given a query actuator also described using the $l$-th pair of normalized features $\mathbf{q}_l\in \mathbb{R}^2$, its distance to the decision hyperplane of sub-SVM$_k$ can be calculated as
\begin{equation}
    d_{lk}(\mathbf{q}_l) = \sum_{i=1}^{N_l}w_{lk}^i y_{lk}^i k(\cfg_{lk}^i, \mathbf{q}_l) -b_{lk},\quad \forall l, k.
\end{equation}
The notations $\cfg_{lk}^i\in \mathbb{R}^2$ and $y_{lk}^i\in \{+1, -1\}$ are overloaded here to simplify symbols used, which denote training samples and labels of each sub-SVM, respectively. They are extracted directly from the whole dataset $(\Cfg', \mathbf{y})$. Classification of the queried actuator $\mathbf{q}_l$ utilizes a max-wins voting strategy, wherein every binary classifier assigns the instance to one of the two classes $(c_{k}^{-}, c_k^+)$ according to the sign of the distance, $\sign(d_{lk}(\mathbf{q}_l))$, 
increases the vote for the assigned class by one, and finally classifies the instance as the class with the most votes.

In summary, $C_5^2\cdot C_7^2 = 210$ sub-SVMs are trained to solve the multi-class classification problem using all pairs of features. For a queried actuator, a class prediction is given for every pair of features available.

\begin{figure*} 
\centering
\vspace{2mm}
\begin{subfigure}{0.19\textwidth}
    \includegraphics[width=\linewidth]{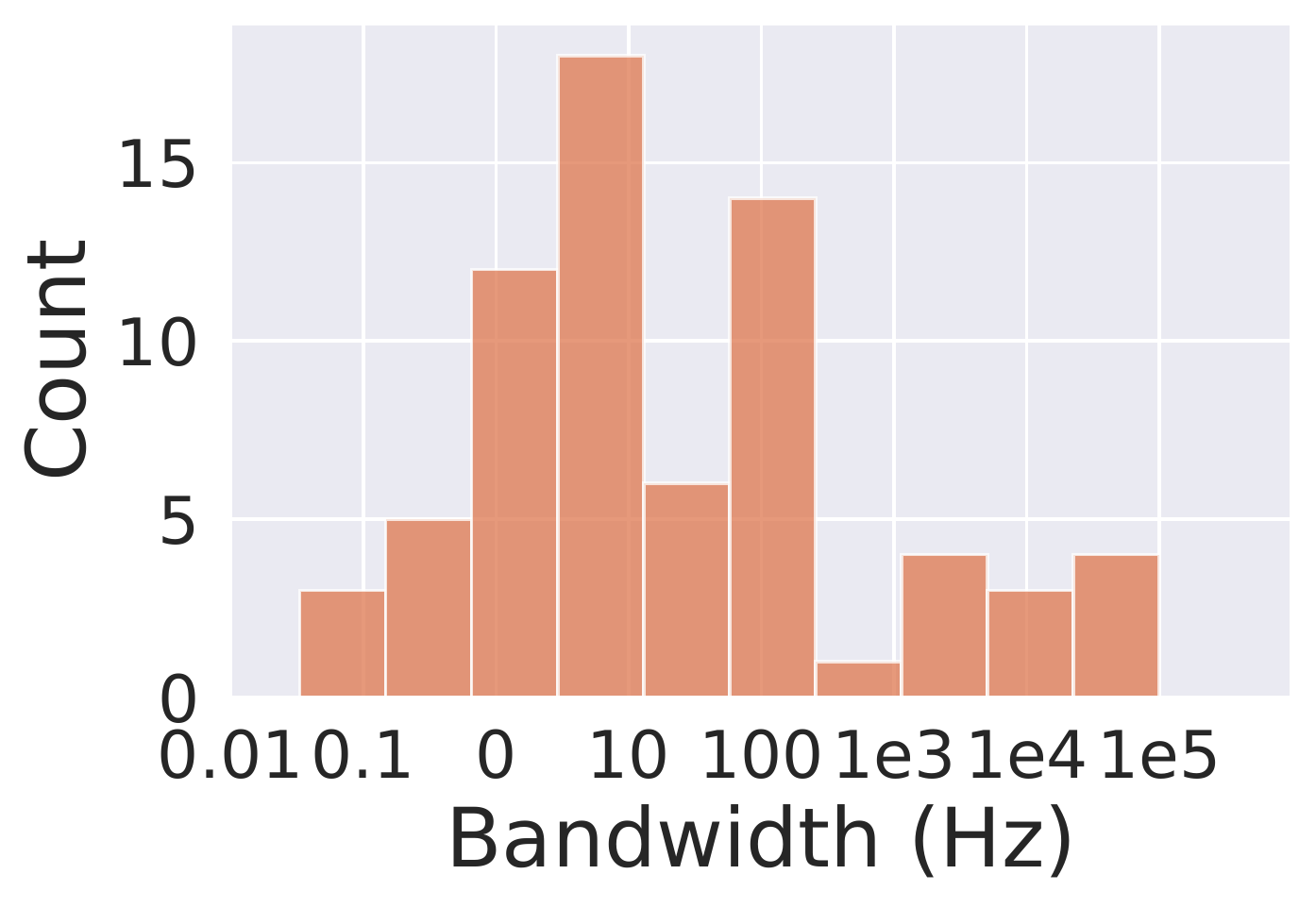}
\end{subfigure}
\begin{subfigure}{0.19\textwidth}
    \includegraphics[width=\linewidth]{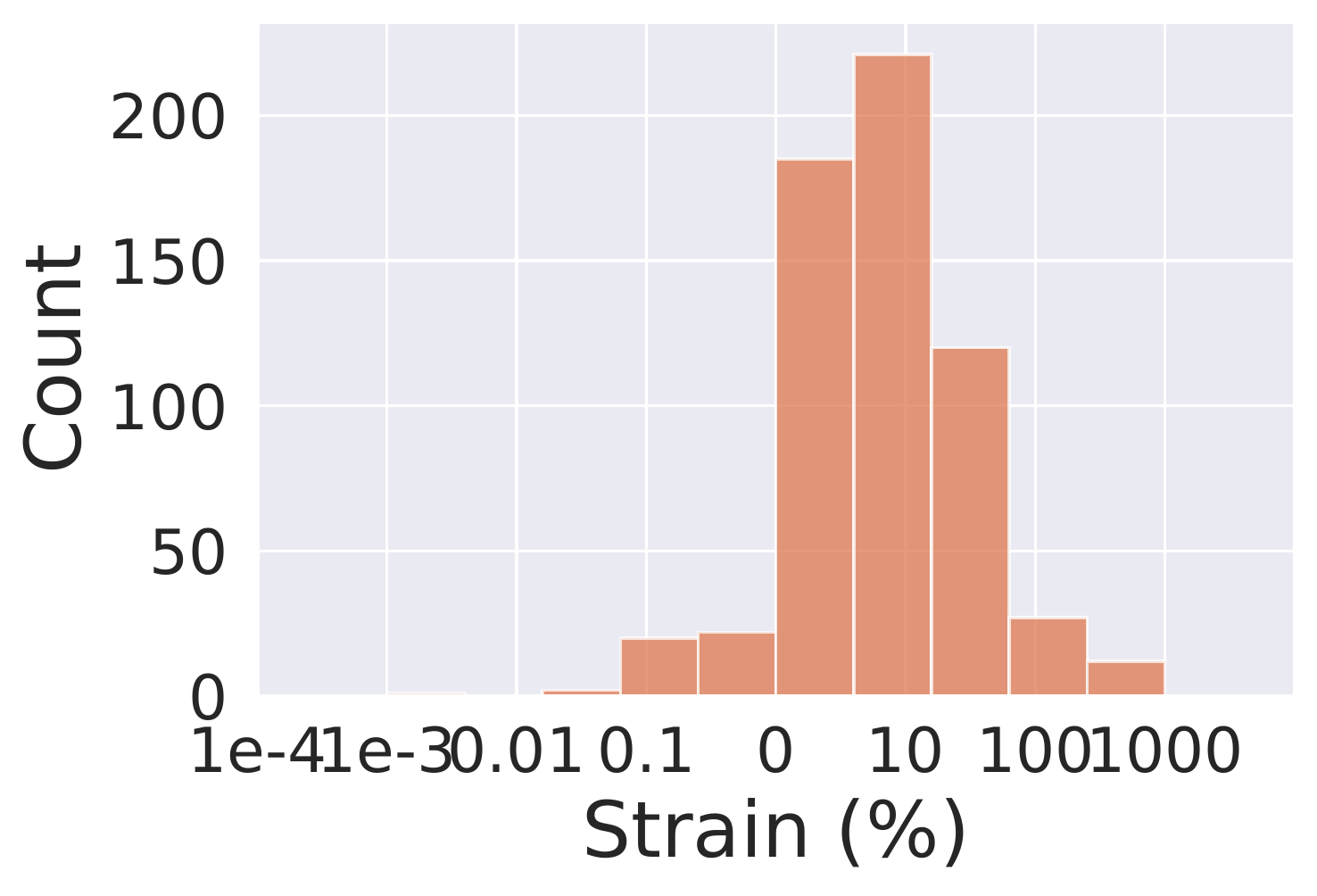}
\end{subfigure}
\begin{subfigure}{0.19\textwidth}
    \includegraphics[width=\linewidth]{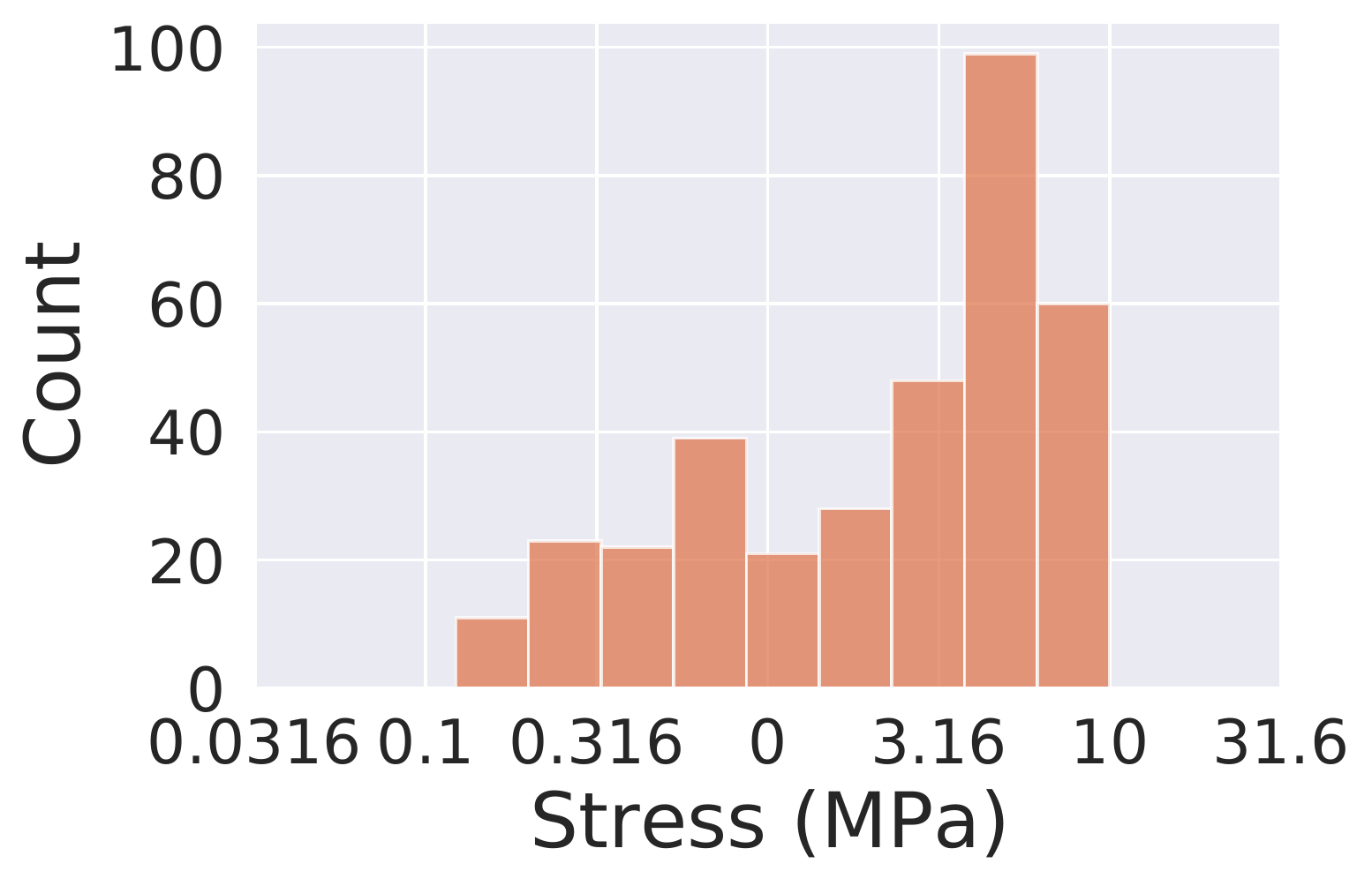}
\end{subfigure}
\begin{subfigure}{0.19\textwidth}
    \includegraphics[width=\linewidth]{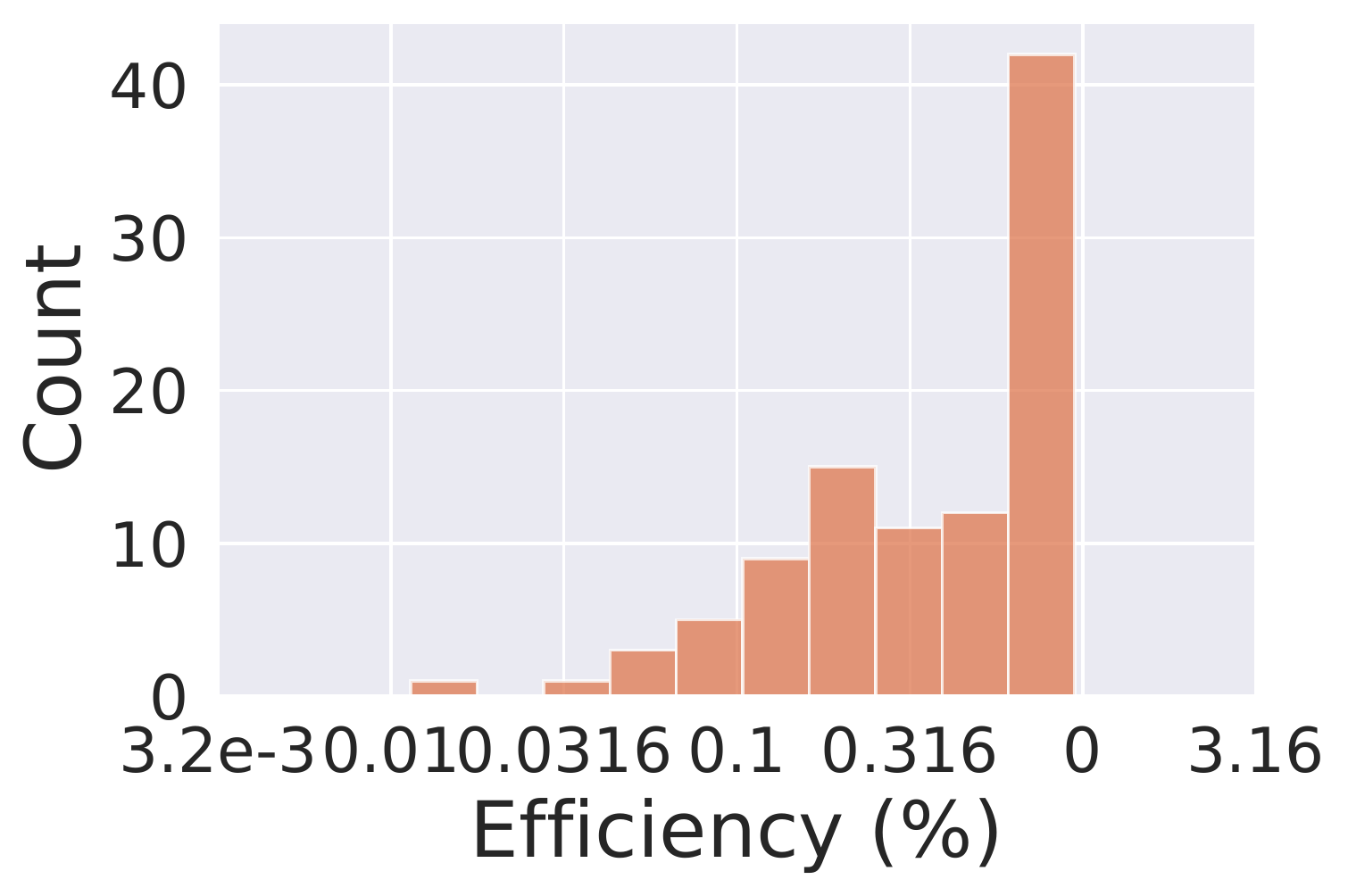}
\end{subfigure}
\begin{subfigure}{0.19\textwidth}
    \includegraphics[width=\linewidth]{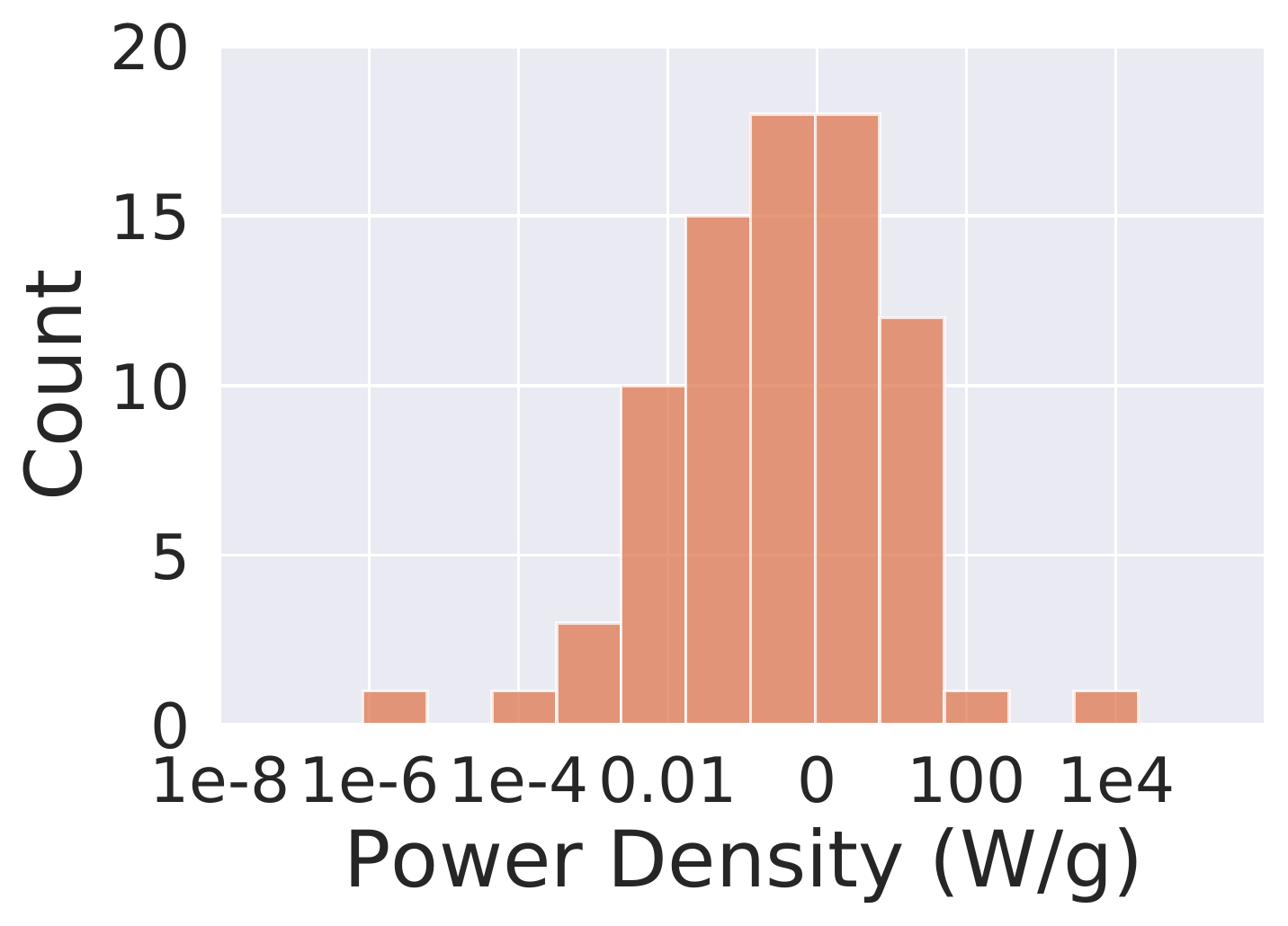}
\end{subfigure}
\vspace{-2mm}
\caption{Histograms pertaining to each actuator feature, displaying the distribution of data points over the range of values. The $\cfg$-axes are in the logarithmic scale.}
\label{histograms}
\vspace{-2mm}
\end{figure*}

Due to the potential overlap of certain actuators across various features, we add the normalized total confidence of each class $j\in\mathcal{C}$ on top of the number of votes to produce a final confidence score,
\begin{align}
    &\begin{aligned}
    D_j(\mathbf{q}_l) &= &\sum_{k=1}^{K} -\mathbf{I}\{c_k^-=j\land d_{lk}\leq0\}d_{lk}(\mathbf{q}_l) \\
    & &+ \mathbf{I}\{c_k^+=j\land d_{lk}>0\}d_{lk}(\mathbf{q}_l),
    \end{aligned} \\
    &S_j(\mathbf{q}_l) = \mathrm{Vote}_j(\mathbf{q}_l) + \frac{D_j(\mathbf{q}_l)}{3|D_j(\mathbf{q}_l)|+1}, \quad \forall j \in \mathcal{C}, \label{eq:confidence_score}
\end{align}
where $\mathbf{I}\{\cdot\}$ equals 1 if the input condition is true and 0 if false. The confidence score allows users to choose among several potential actuator muscle options for their needs while knowing the relative chances of success given their application constraints. 
The confidence score $S$ can also be used to break ties when some classes get equal votes. Since the added part
\begin{equation}
\frac{D_j(\mathbf{q}_l)}{3\left|D_j(\mathbf{q}_l)\right|+1} \in \left(-\frac13, \frac13\right),    
\end{equation}
it preserves the original order of classes by the number of votes except breaking ties among equally voted classes.

It is important to mention an alternative method of calculating multi-class confidence scores for SVMs. It produces probabilistic estimations through a combination of the Platt scaling algorithm \cite{platt1999probabilistic} and pairwise coupling \cite{wu2004probability}. We will show in experiments that, however, due to the collected dataset not being well balanced among classes and the relatively small number of training samples, the probabilistic estimation method provides less accurate predictions than the selected confidence score.

\begin{table} 
\caption{Metrics of the collected muscle actuator dataset. This table gives a high-level overview of the 678 data points collected. It shows that by taking logarithm the variances of features are reduced by a large margin and their distributions are made more uniform, thus numerically easier to handle by machine learning algorithms.}
\centering
\resizebox{\linewidth}{!}{%
\begin{tabular}{cccccc}\toprule
 & Count & Range &  Mean & Var. & Log Var. \\
 \midrule
 Bandwidth (Hz) & 73 & (0.0333, 100000.0) & 4658.63 & 3.2903e+08 & 2.2347\\
 Strain (\%) & 619 & (0.001, 1000.0) & 24.5232 & 4819.03 & 0.5012\\
 Stress (MPa) & 355 & (0.15, 1000.0) & 165.352 & 32420.4 & 0.9676\\
 Efficiency (\%) & 100 & (0.013, 90.0) & 38.6423 & 1377.13 & 0.6872\\
 Power Density (W/g) & 81 & (8.23e-07, 20000.0) & 252.93 & 4.9355e+06 & 2.3777\\
 \bottomrule
\end{tabular}
\label{data_statistics}
}
\vspace{-4mm}
\end{table}

\section{Results}

\subsection{Statistics and Distribution of Data}\label{sec:res_data}
Our resulting training dataset includes 678 data points across 7 types of actuators and 5 parameters as described in Sec.~\ref{sec:data}.
Table \ref{data_statistics} provides a high-level overview of the collected dataset with relevant metrics. We additionally provide histograms (Fig. \ref{histograms}) for each actuator feature to display the distribution of data points over the range of values.



\subsection{Comparing Hyperparameters and Decision Methods}

To determine the best hyperparameters for training SVMs, 5-fold cross-validations are run on the dataset. The most important hyperparameter of an SVM is the regularization term $\lambda$. The metrics considered are top-1 and top-3 accuracy averaged by class labels (macro average). Macro average is selected in case the actuator with the most data samples dominates the metric value.
When calculating the confidence for various classes of actuators given a set of necessary features, it is important to determine whether to use the confidence score mentioned in Eq.~\eqref{eq:confidence_score} or the probabilistic estimation formulated by pairwise coupling \cite{wu2004probability}. 

\begin{figure}[tbp]
    \centering
    \begin{subfigure}{.85\linewidth}
        \includegraphics[width=0.99\linewidth]{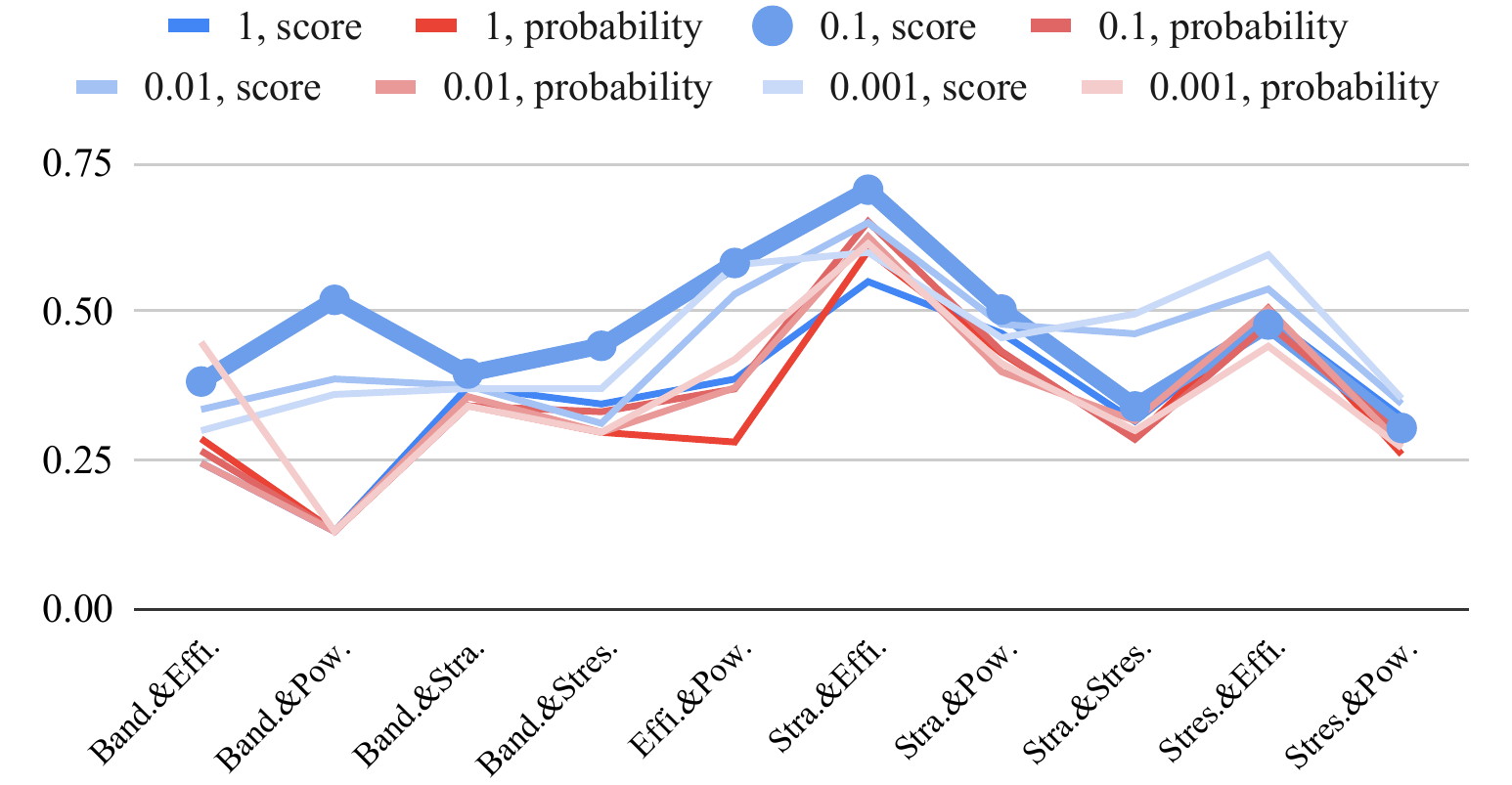}
        \caption{Macro average top-1 acc.}
    \end{subfigure}
    \begin{subfigure}{.85\linewidth}
        \includegraphics[width=0.99\linewidth]{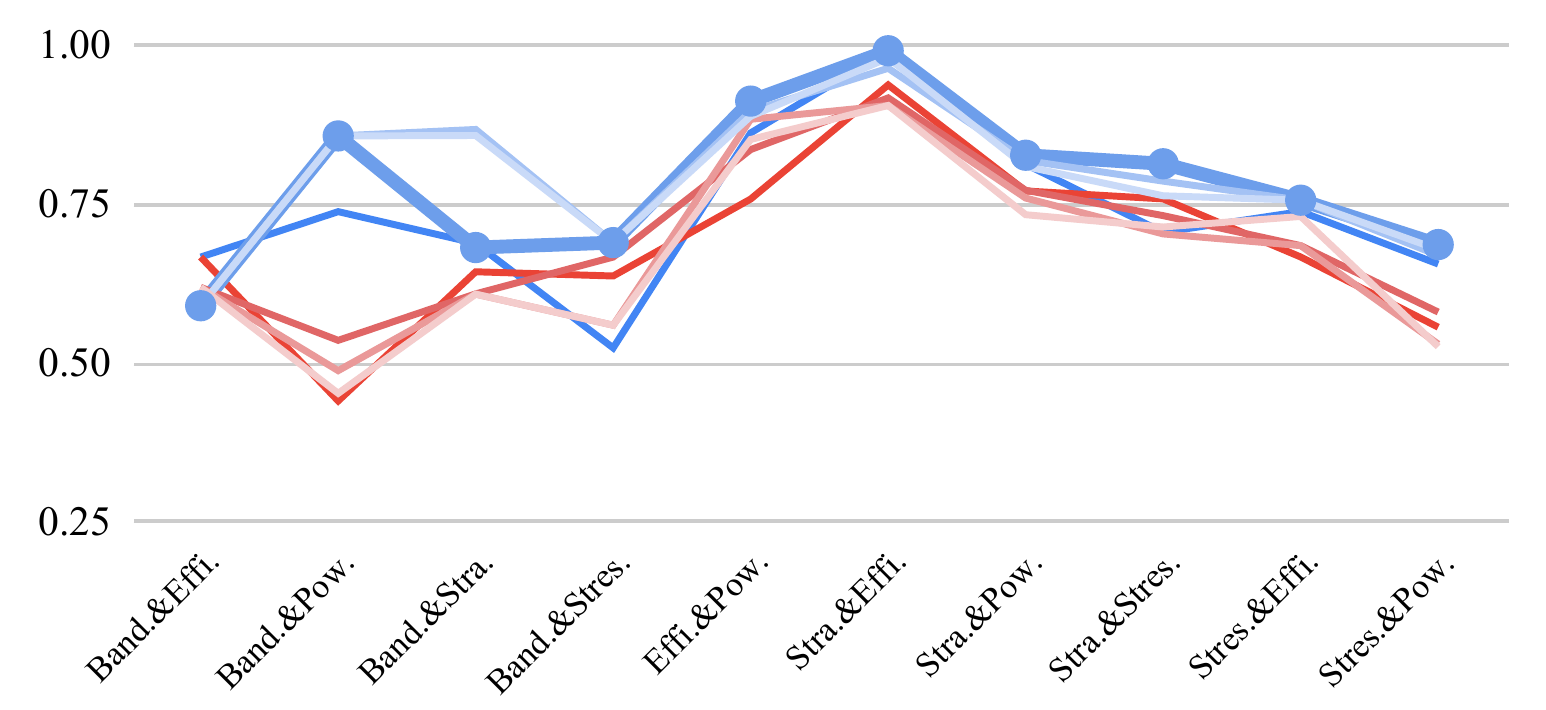}
        \caption{Macro average top-3 acc.}
    \end{subfigure}
    
    \caption{Comparing different hyperparameters and decision methods for the SVMs. The $\mathbf{x}$-axes lists different pairs of features used, and the $\mathbf{y}$-axis is accuracy values averaged among classes. Blue lines use the confidence score in Eq.~\eqref{eq:confidence_score}, red lines use probability. Different shades of lines denote the value of $\lambda$. It is shown the confidence score is in general a more suitable decision method on this dataset, and $\lambda=0.1$ seems to be the best regularization value overall. The lines generated by the chosen hyperparameter and decision method are highlighted by dot markers and wider strokes. }
    \label{fig:acc}
    \vspace{-4mm}
\end{figure}

We combine these two sets of experiments, i.e., cross-validate SVMs with every possible combination of $\lambda \in \{1, 0.1, 0.01, 0.001\}$ and use the score or probability. The results are shown in Fig.~\ref{fig:acc}, which shows the score value defined by Eq.~\eqref{eq:confidence_score} produces more accurate predictions in general than probability, and $\lambda=0.1$ seems to be an overall best value for regularization.

\subsection{SVM Classification of the Performance Space}

\begin{figure*}[t]
\vspace{3mm}
\begin{subfigure}{0.19\textwidth}
    \includegraphics[width=\linewidth, trim={8mm 7mm 2.5mm 2mm}, clip=true]{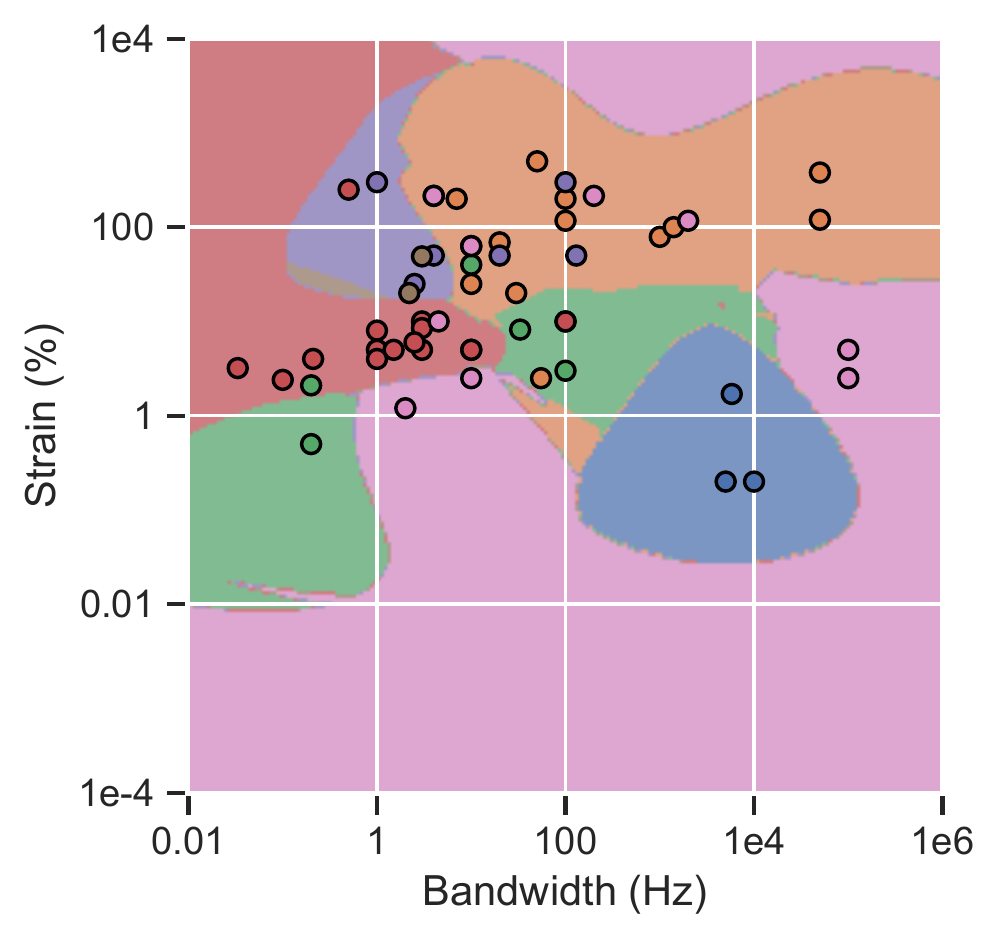}
    \vspace{-5mm}
    \caption{\tiny Strain v. Bandwidth, 56 pts}
    \label{multiclass_contour_a}
    \vspace{2mm}
\end{subfigure}
\begin{subfigure}{0.19\textwidth}
    \includegraphics[width=\linewidth, trim={8mm 7mm 2.5mm 2mm}, clip=true]{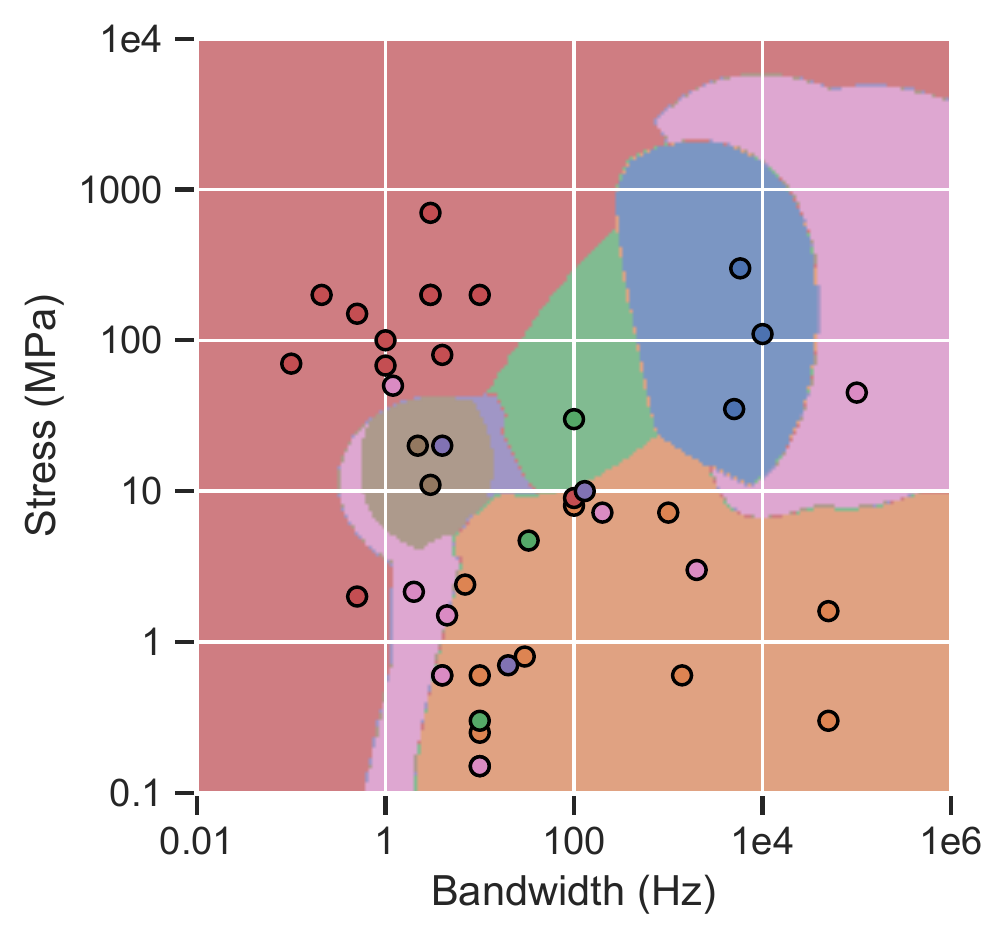}
    \vspace{-5mm}
    \caption{\tiny Stress v. Bandwidth, 42 pts}
    \label{multiclass_contour_b}
    \vspace{2mm}
\end{subfigure}
\begin{subfigure}{0.19\textwidth}
    \includegraphics[width=\linewidth, trim={8mm 7mm 2.5mm 2mm}, clip=true]{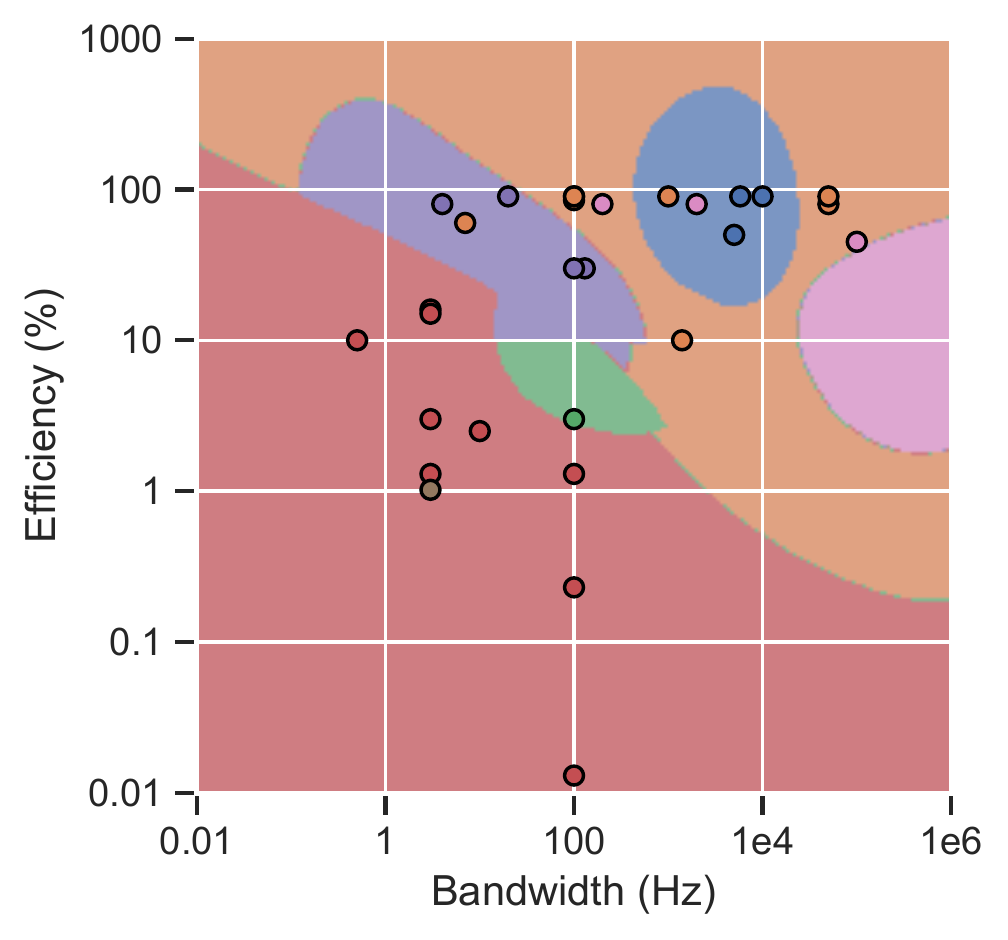}
    \vspace{-5mm}
    \caption{\tiny Effi. v. Bandwidth, 29 pts}
     \label{multiclass_contour_c}
     \vspace{2mm}
\end{subfigure}
\begin{subfigure}{0.19\textwidth}
    \includegraphics[width=\linewidth, trim={8mm 7mm 2.5mm 2mm}, clip=true]{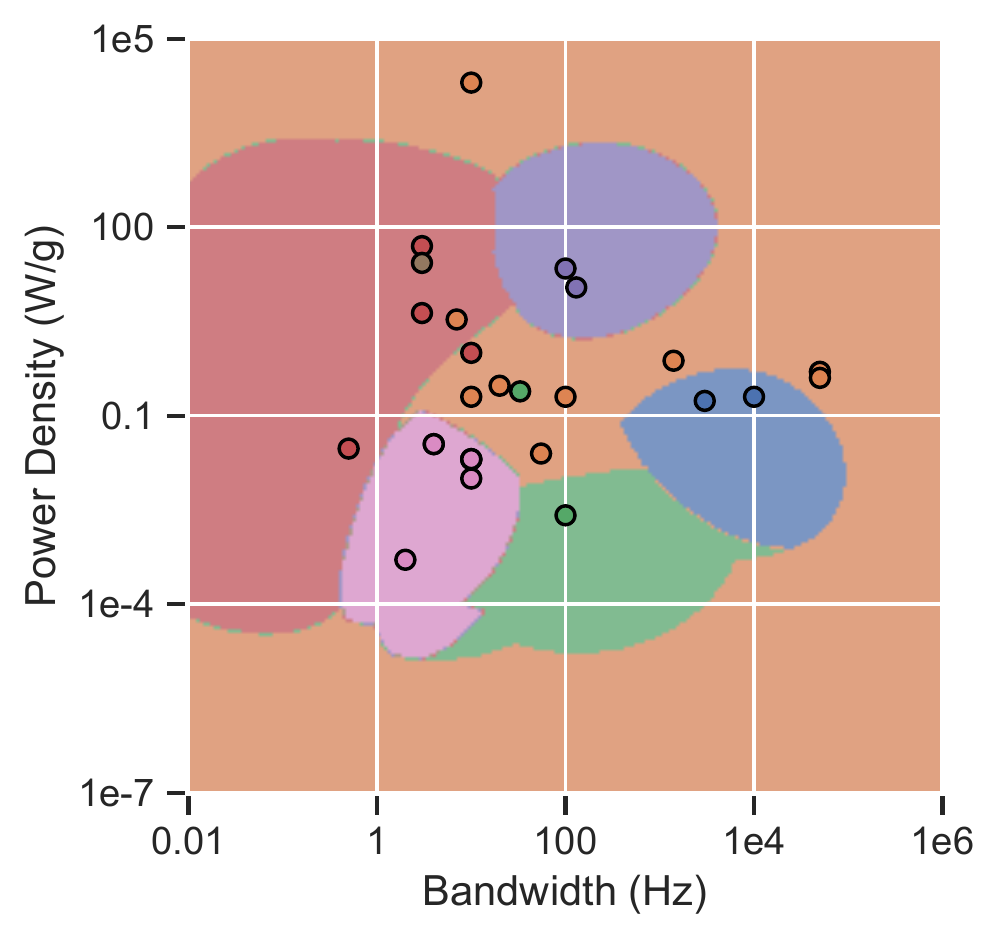}
    \vspace{-5mm}
    \caption{\tiny Pow. v. Bandwidth, 27 pts}
    \vspace{2mm}
     \label{multiclass_contour_d}
\end{subfigure}
\begin{subfigure}{0.19\textwidth}
    \includegraphics[width=\linewidth, trim={8mm 7mm 2.5mm 2mm}, clip=true]{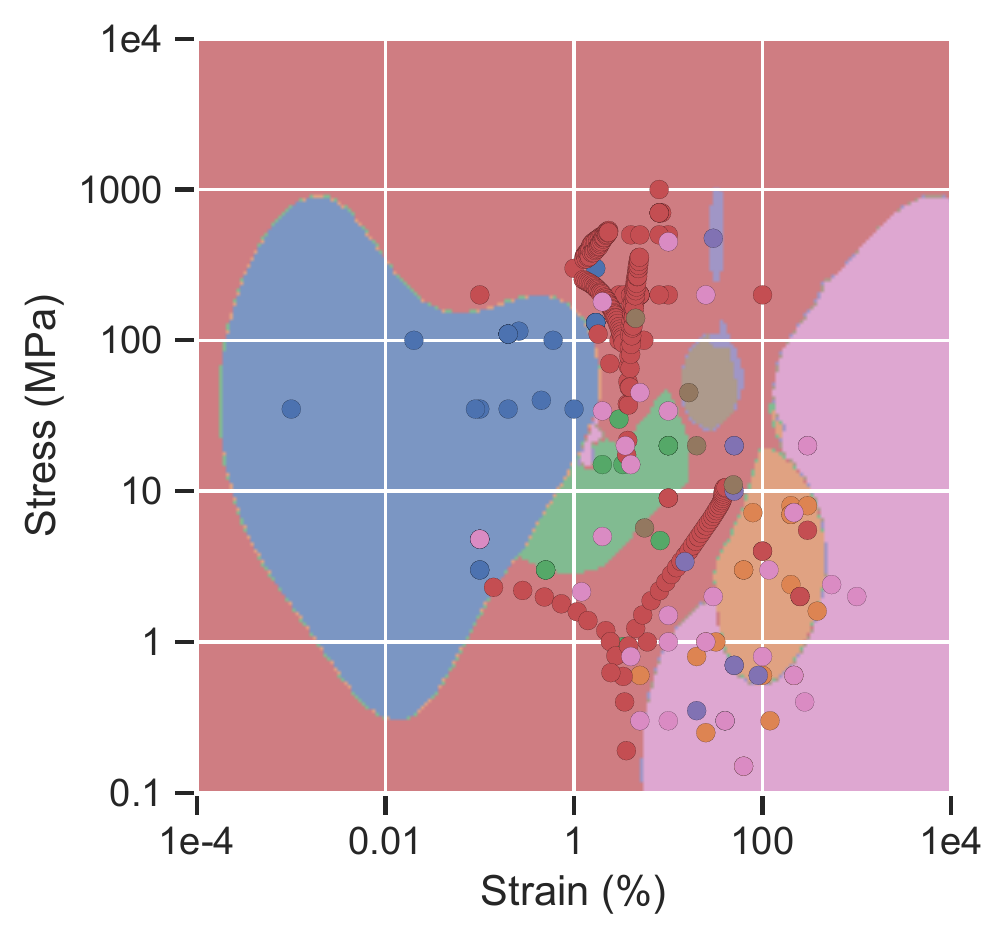}
    \vspace{-5mm}
    \caption{\tiny Stress v. Strain, 343 pts}
    \vspace{2mm}
     \label{multiclass_contour_e}
\end{subfigure}
\begin{subfigure}{0.19\textwidth}
    \includegraphics[width=\linewidth, trim={8mm 7mm 2.5mm 2mm}, clip=true]{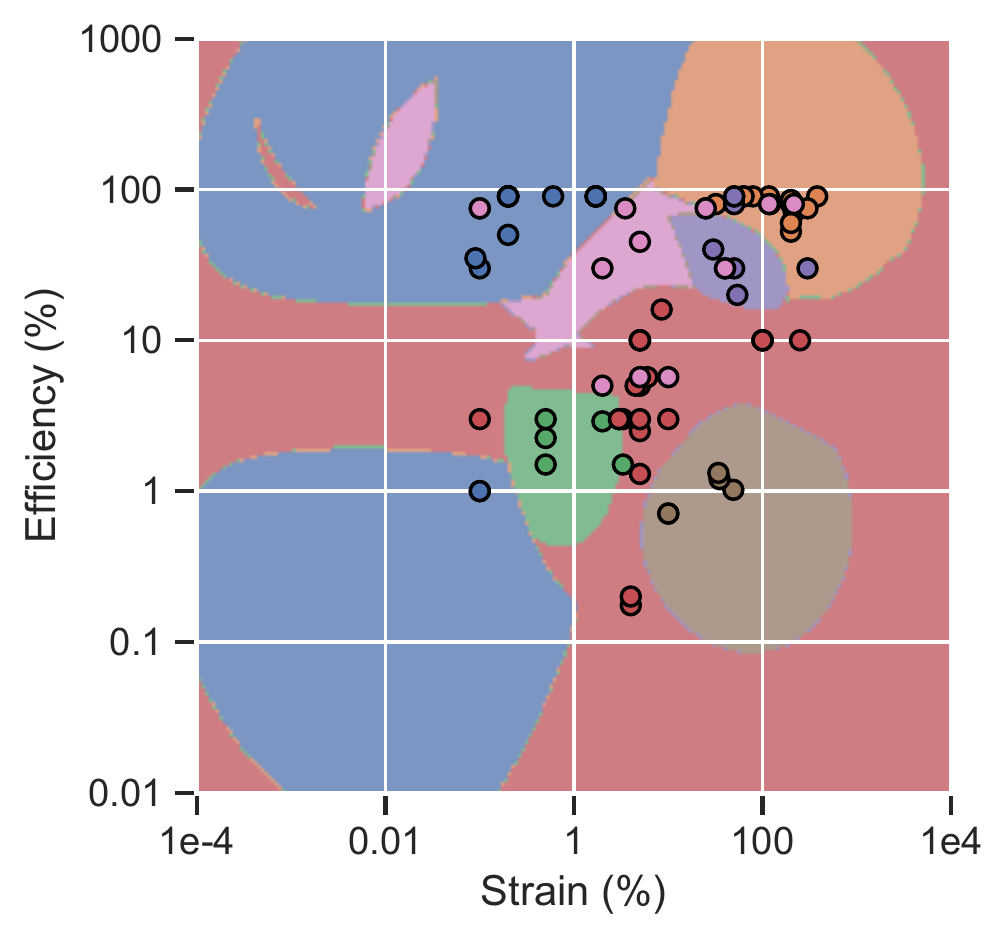}
    \vspace{-5mm}
    \caption{\tiny Efficiency v. Strain, 86 pts}
     \label{multiclass_contour_f}
\end{subfigure}
\begin{subfigure}{0.19\textwidth}
    \includegraphics[width=\linewidth, trim={8mm 7mm 2.5mm 2mm}, clip=true]{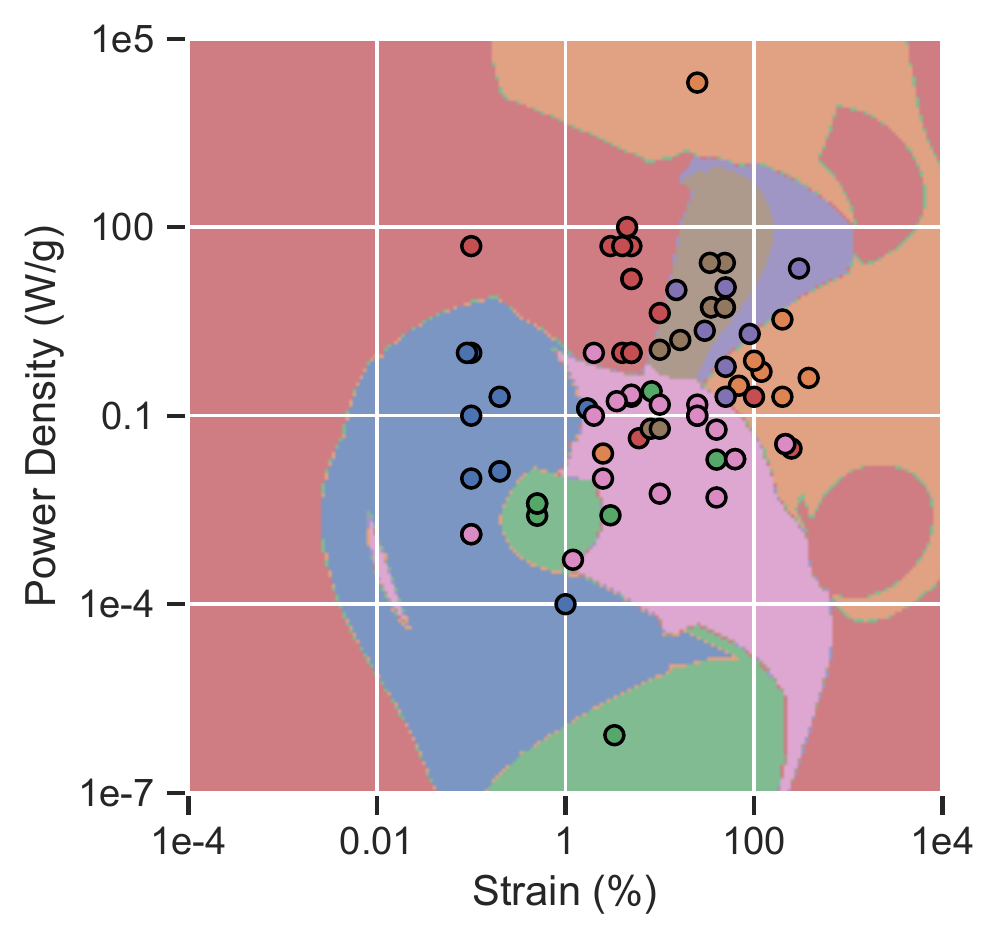}
    \vspace{-5mm}
    \caption{\tiny Pow Density v. Strain, 77 pts}
     \label{multiclass_contour_g}
\end{subfigure}
\begin{subfigure}{0.19\textwidth}
    \includegraphics[width=\linewidth, trim={8mm 7mm 2.5mm 2mm}, clip=true]{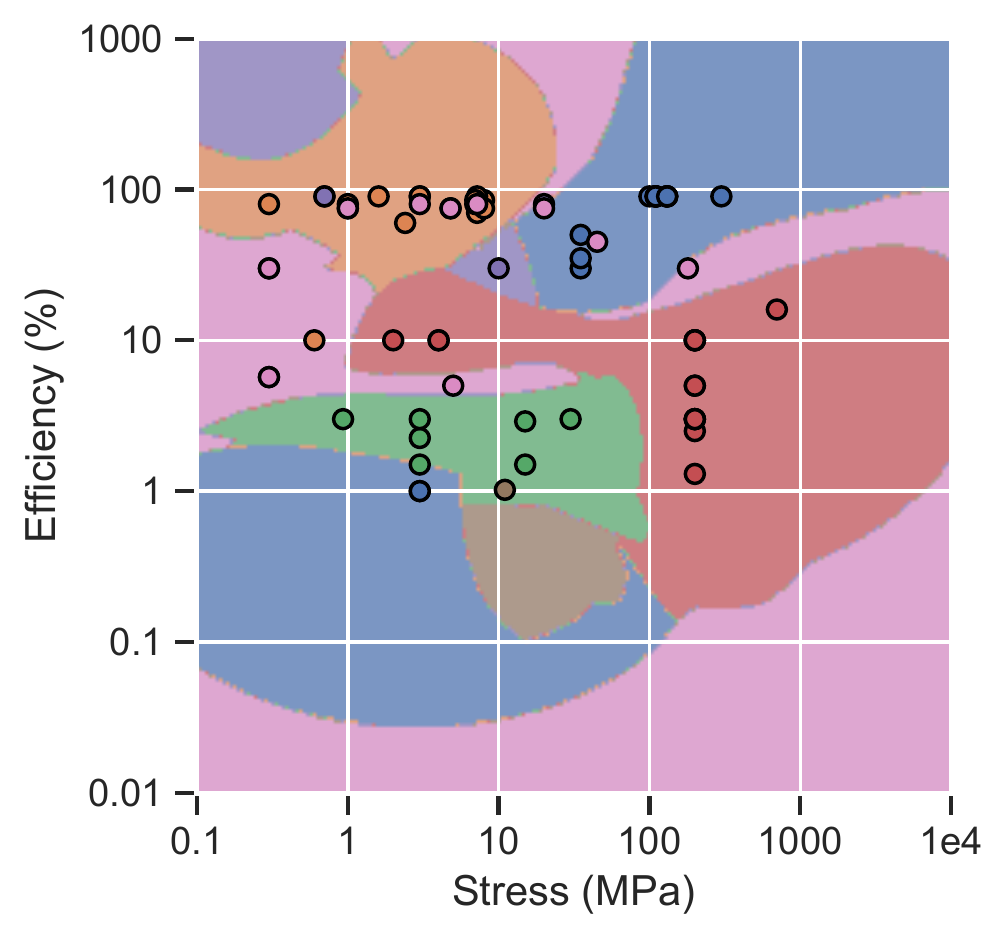}
    \vspace{-5mm}
    \caption{\tiny Efficiency v. Stress, 73 pts}
     \label{multiclass_contour_h}
\end{subfigure}
\begin{subfigure}{0.19\textwidth}
    \includegraphics[width=\linewidth, trim={8mm 7mm 2.5mm 2mm}, clip=true]{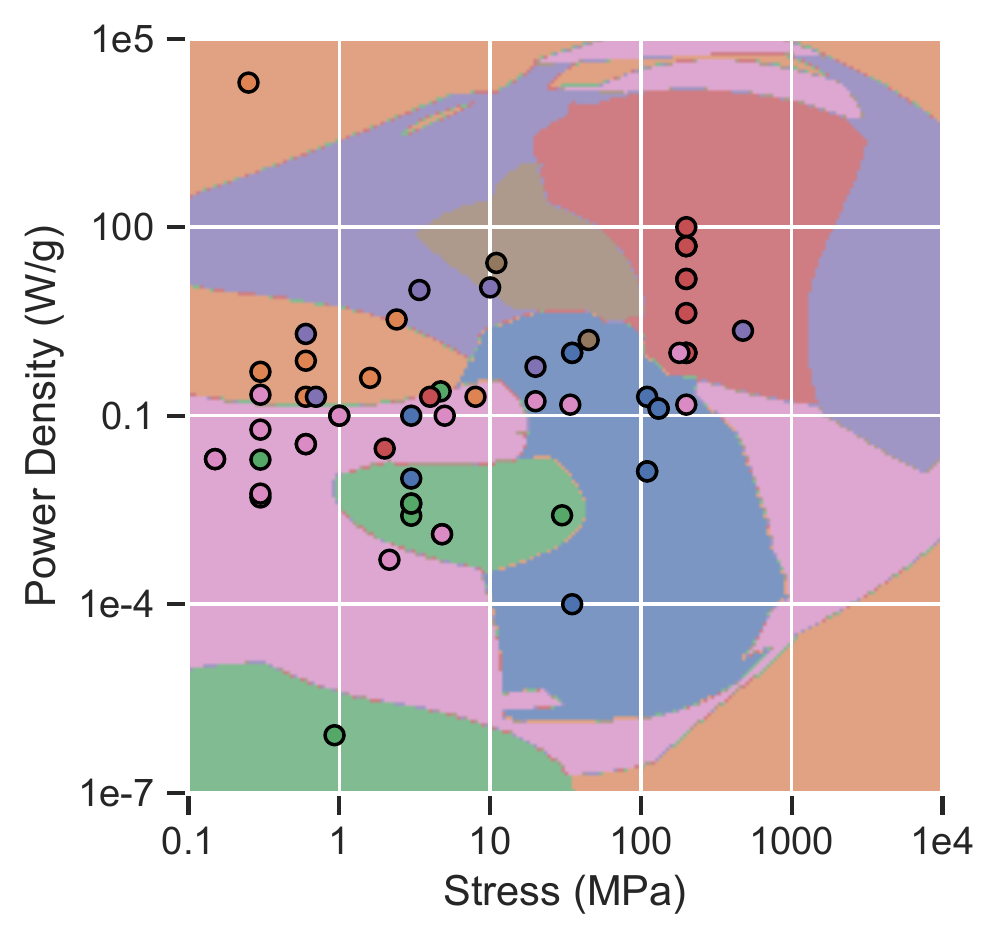}
    \vspace{-5mm}
    \caption{\tiny Pow Density v. Stress, 62 pts}
     \label{multiclass_contour_i}
\end{subfigure}
\begin{subfigure}{0.19\textwidth}
    \includegraphics[width=\linewidth, trim={8mm 7mm 2.5mm 2mm}, clip=true]{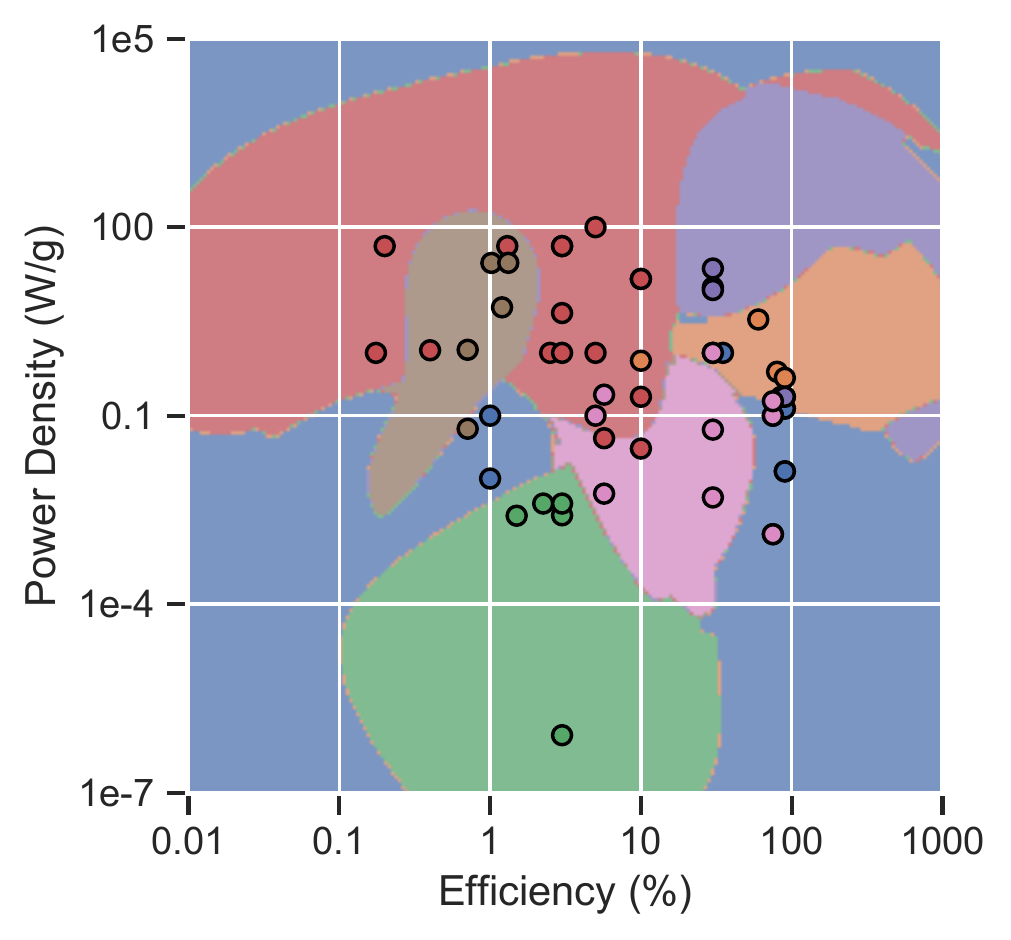}
    \vspace{-5mm}
    \caption{\tiny Pow. v. Efficiency, 55 pts}
     \label{multiclass_contour_j}
\end{subfigure}
\begin{subfigure}{0.4\textwidth}
    \centering
    \vspace{1mm}
    \includegraphics[width=\linewidth]{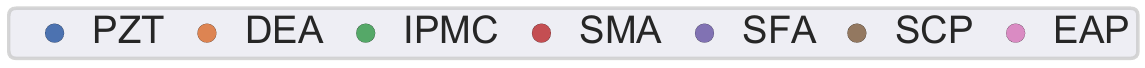} 
     \label{fig:legend}
\end{subfigure}
\centering
\vspace{-6mm}
\caption{Multi-class contours for bi-variate classifiers across all 5 actuator features. 
The colors of the data points indicate their actuator class, and the corresponding color fill on each plot corresponds to the SVM regions for each class. A test point that lies within one of these regions will be classified as such. The figures are read as y-axis v. x-axis.}
\label{multiclass_contour}
\end{figure*}

\begin{table*}[ht]
\centering
\caption{Top-1 and top-3 accuracies for SVMs using different pairs of features. The highlighted classifiers are the most accurate ones for each individual class of actuators. Users of the proposed tool should refer to this table about what features to consider when deciding the actuator type of robot muscle applications. Most 0's in this table exist because of absence of or low number of valid samples belonging to the class for that pair of features.}
\label{tab:acc}
\resizebox{\textwidth}{!}{%
\begin{tabular}{lccccccccc  ccccccccc}
\toprule
\multirow{2}{*}{Features} & \multicolumn{9}{c}{Top-1}                                                                                                                                 & \multicolumn{9}{c}{Top-3}                                                                                                                                     \\ \cmidrule(l){2-10} \cmidrule(l){11-19}
                          & PZT             & DEA            & IPMC           & SMA            & SFA             & SCP             & EAP            & Macro         & Micro     & PZT             & DEA             & IPMC            & SMA             & SFA             & SCP             & EAP             & Macro     & Micro     \\ \midrule
Band. \& Effi.              & 75.00           & 28.57          & 0.00           & 88.89          & 75.00           & 0.00            & 0.00           & 38.21          & 55.17          & 75.00           & 71.43           & 0.00            & \textbf{100.00} & \textbf{100.00} & 0.00            & 66.67           & 59.01          & 79.31          \\
Band. \& Pow.               & \textbf{100.00} & 63.64          & 0.00           & 50.00          & \textbf{100.00} & 0.00            & 50.00          & 51.95          & 55.56          & \textbf{100.00} & \textbf{100.00} & \textbf{100.00} & 100.00          & \textbf{100.00} & 0.00            & \textbf{100.00} & 85.71          & 96.30          \\
Band. \& Stra.              & 66.67           & 80.00          & 14.29          & 71.43          & 0.00            & 0.00            & 44.44          & 39.55          & 51.79          & \textbf{100.00} & 86.67           & 71.43           & 85.71           & 66.67           & 0.00            & 66.67           & 68.16          & 76.79          \\
Band. \& Stres.             & \textbf{100.00} & 63.64          & 0.00           & 83.33          & 0.00            & 50.00           & 12.50          & 44.21          & 52.38          & \textbf{100.00} & 90.91           & 33.33           & 83.33           & 0.00            & \textbf{100.00} & 75.00           & 68.94          & 76.19          \\
Effi. \& Pow.               & 20.00           & 50.00          & \textbf{80.00} & 68.75          & 75.00           & 80.00           & 33.33          & 58.15          & 54.55          & \textbf{100.00} & 83.33           & 80.00           & \textbf{100.00} & 75.00           & \textbf{100.00} & \textbf{100.00} & 91.19          & 94.55          \\
Stra. \& Effi.              & \textbf{100.00} & \textbf{87.50} & 71.43          & 75.00          & 42.86           & \textbf{100.00} & 16.67          & \textbf{70.49} & 73.26          & \textbf{100.00} & 93.75           & \textbf{100.00} & \textbf{100.00} & \textbf{100.00} & \textbf{100.00} & \textbf{100.00} & \textbf{99.11} & \textbf{98.84} \\
Stra. \& Pow.               & 63.64           & 33.33          & 28.57          & 73.33          & 0.00            & 80.00           & \textbf{73.33} & 50.32          & 55.84          & 90.91           & 75.00           & 57.14           & 86.67           & 85.71           & 90.00           & 93.33           & 82.68          & 84.42          \\
Stra. \& Stres.             & 65.38           & 57.89          & 7.69           & \textbf{93.78} & 0.00            & 0.00            & 13.33          & 34.01          & \textbf{75.51} & \textbf{100.00} & 94.74           & 84.62           & 100.00          & 66.67           & 40.00           & 83.33           & 81.34          & 95.92          \\
Stres. \& Effi.             & 78.95           & 85.71          & 42.86          & 93.75          & 0.00            & 0.00            & 33.33          & 47.80          & 67.12          & 94.74           & 92.86           & \textbf{100.00} & \textbf{100.00} & 50.00           & 0.00            & 91.67           & 75.61          & 91.78          \\
Stres. \& Pow.              & 63.64           & 30.00          & 14.29          & 83.33          & 0.00            & 0.00            & 21.43          & 30.38          & 38.71          & 90.91           & 80.00           & 57.14           & 83.33           & 33.33           & 50.00           & 85.71           & 68.63          & 75.81          \\ \bottomrule
\end{tabular}%
}
\vspace{-2mm}
\end{table*}

Using the hyperparameter and decision method determined in the last experiment, we train SVMs across each pair of features. This section first visually analyses the performance space defined by the SVMs, then presents the 5-fold cross-validation results of these models. The models for visualization purposes are trained on all valid samples, and are used in our open-source platform.

Since we train a multi-class classifier for each pair of features, we obtain 10 separate classifiers.
With these models, we can predict which type of actuator will work best when given a certain set of parameters. 
Fig.~\ref{multiclass_contour} visualizes these bi-variate classifiers. We can see that while some actuator types are almost linearly separable, many of them have severe overlap that result in complicated classifier boundary regions. For example, in Fig. \ref{multiclass_contour_c} depicting efficiency v. bandwidth, we see a distinct separation between both SMA and SCP actuators and the remaining classes. However, these two classes share similar properties, and thus their distributions tend to overlap or lie near one another in each classifier. Similarly, DEA and EAP actuators tend to have analogous features, so it is no surprise that their data points in these models tend to lie close to one another. In these results emerge several muscle groupings, where distinct actuator technologies share very similar properties and performance ranges. This provides valuable insight into the space of actuators that are valid options for a particular application given certain parameter constraints.

From Fig. \ref{multiclass_contour} we can also see that the data tends to be clustered and the boundaries between predicted actuator distributions are noisy. Because the different actuator types have features with varying orders of magnitude, logarithmic normalization must be performed. Unfortunately, this results in data points collected becoming very clustered. 
This can lead to errors in the SVMs trained and result in complicated and overlapping distributions. 

\begin{table*}[t]
\centering
\vspace{3mm}
\caption{Top-4 predictions and associated confidence scores generated by different classifiers using available features of 5 example robot muscles. The correct predictions are highlighted. This suggests the proposed method is able to yield reasonable predictions of actuator type to use in real-world applications given the needed constraints. The available features are given in the order of \{Bandwidth (Hz), Strain (\%), Stress (MPa), Efficiency (\%), Power Density (W/g)\}.}
\label{tab:app}
\resizebox{.95\textwidth}{!}{%
\begin{tabular}{@{}lccccccccc} 
\toprule
Application                                                                         & \multicolumn{6}{c}{Artificial Bicep on Skeletal Arm \cite{bicep} \{N/A, 25, 1.0, 75, 0.1\}}                                                                                    & \multicolumn{3}{c}{Microrobot \cite{microrobot} \{200, 2.5, N/A, N/A, 0.3\}}   \\ \cmidrule{1-1} \cmidrule(l){2-7} \cmidrule(l){8-10}
Features                                                                            & Stra./Stres.                      & Stra./Effi.                       & Stra./Pow.            & Stres./Effi.           & Stres./Pow.           & Effi./Pow.           & Band./Stra.           & Band./Pow.            & Stra./Pow.          \\ \midrule
\multirow{4}{*}{\begin{tabular}[c]{@{}l@{}}Predictions\\ (Confidence)\end{tabular}} & EAP (6.3)                         & \textbf{DEA (6.3)}                & EAP (6.3)             & \textbf{DEA (6.3)}     & EAP (6.3)             & PZT (6.3)            & IPMC (6.3)            & DEA (6.3)             & EAP (6.3)           \\
                                                                                    & \textbf{DEA (5.3)}                & SFA (5.3)                         & \textbf{DEA (5.3)}    & EAP (5.3)              & \textbf{DEA (5.3)}    & \textbf{DEA (5.3)}   & DEA (5.3)             & IPMC (5.3)            & \textbf{PZT (5.2)}  \\
                                                                                    & SFA (4.3)                         & EAP (4.3)                         & SCP (3.2)             & SFA (4.3)              & SMA (3.2)             & EAP (4.3)            & SMA (4.2)             & \textbf{PZT (3.8)}    & SMA (4.3)           \\
                                                                                    & SMA (3.2)                         & SMA (2.9)                         & IPMC (3.1)            & PZT (2.8)              & SFA (3.1)             & SFA (2.8)            & \textbf{PZT (2.9)}    & SMA (2.8)             & DEA (3.2)           \\ \bottomrule\toprule
Application                                                                         & \multicolumn{2}{c}{Active Endoscope\cite{endoscope} \{0.08, 5\}} & \multicolumn{3}{c}{Rehabilitative Glove \cite{glove} \{0.5, 5, 0.35\}} & \multicolumn{3}{c}{Artificial Finger \cite{finger} \{0.2, 10, 0.3\}} &                     \\ \cmidrule{1-1} \cmidrule(l){2-3} \cmidrule(l){4-6}  \cmidrule(l){7-9}
Features                                                                            & \multicolumn{2}{c}{Band./Stra.}                                       & Band./Stra.           & Band./Stres.           & Stra./Stres.          & Band./Stra.          & Band./Stres.          & Stra./Stres.          &                     \\ \midrule
\multirow{4}{*}{\begin{tabular}[c]{@{}l@{}}Predictions\\ (Confidence)\end{tabular}} & \multicolumn{2}{c}{\textbf{SMA (6.3)}}                                & SMA (6.3)             & SMA (6.3)              & SMA (6.3)             & SMA (6.3)            & SMA (6.3)             & EAP (6.3)             &                     \\
                                                                                    & \multicolumn{2}{c}{IPMC (5.3)}                                        & IPMC (5.3)            & EAP (5.3)              & EAP (5.3)    & \textbf{SCP (5.3)}   & EAP (5.3)             & DEA (5.3)             &                     \\
                                                                                    & \multicolumn{2}{c}{EAP (4.1)}                                         & EAP (4.3)             & DEA (4.2)              & DEA (4.2)             & SFA (3.2)            & DEA (4.2)             & SFA (4.2)             &                     \\
                                                                                    & \multicolumn{2}{c}{SCP (2.8)}                                         & SCP (2.8)             & IPMC (2.8)    & \textbf{SFA (2.9)}    & IPMC (3.2)           & IPMC (2.8)            & SMA (3.2)             &                     \\ \bottomrule
\end{tabular}%
}
\vspace{-5mm}
\end{table*}

Thus, in addition to providing a top-1 prediction of the ideal actuator class for a set of test parameters, we include the top-$n$ actuator type predictions, 
as well as the associated confidence score for each. Table \ref{tab:acc} depicts top-1 and top-3 accuracies of SVMs using different pairs of features. It presents accuracies in each individual class of actuators as well as the accuracy averaged by class (macro) and by sample (micro). For each class, the SVM with the highest accuracy is highlighted. It shows that there exists a reasonably accurate classifier for each class of actuators. Also, it proves that top-3 predictions cover the correct selections in most scenarios if a proper classifier is used, which greatly improves the usability of the proposed robot muscle selection tool. This table also serves as a guide to future users of this tool about what features to consider. 

\subsection{Real-world Applications}
In addition to quantifying the performance of our model on a testing set from our collected data, we also evaluate its success using constraints from several real-world robotic applications. We collect 5 example robotic muscles, and let the SVMs predict their actuator types using each pair of available features/constraints. The top-4 predictions along with the corresponding confidence scores are in Table~\ref{tab:app}. This table shows the correct actuator types can be found by the classifiers among top predictions with relatively high confidence, suggesting its usability in real-world applications. 
In the cases of the Rehabilitative Glove and Artificial Finger, too few samples of SFA and SCP were available in the dataset, making it hard for SVMs to learn about them. It also suggests that SMA and EAP actuators may be good alternatives for these two applications given the required features.




\subsection{Open-access Web Toolkit}
To make the collected robot muscle dataset and the data-driven actuator selection algorithm accessible to the community, we created \textit{Robot Muscle Toolkit}\footnote{\href{http://robotmuscletoolkit.herokuapp.com/research}{http://robotmuscletoolkit.herokuapp.com/research}}, an open-access web application to help people determine the proper category of actuators for their artificial muscle applications.
With the web platform, users can visualize the collected dataset and interactively input the necessary performance parameters of their own applications to see the most likely actuator types. All predictions come with a confidence score to help users decide. The queried features are plotted alongside the training samples to help users intuitively know how similar their needs are to off-the-shelf artificial muscle designs. 






\section{Discussion and Conclusion}
For this paper, we constructed a data-driven approach for actuator selection in artificial muscle applications and delivered an open-access web toolkit utilizing this method.  
This first-of-its-kind method can help users in several critical scenarios.
First, it provides a way to 
determine which actuators are most relevant for a specific range of desired artificial muscle parameters. Confidence scores of potential actuator technologies for achieving this set of performance specifications are provided to inform decisions. 
Secondly, it serves as a unifying benchmark for comparison and performance evaluation of various artificial muscle types. It is able to visually describe the multi-dimensional margins among different muscle groups in performance parameters.
Finally, it can evaluate new and yet-to-be-discovered robot muscle technologies and compare them against the distributions of existing ones.

As part of this effort, we collected hundreds of data points; however, we acknowledge that improving the dataset and the way we use it may very well improve the performance of the model. 
A major drawback of our current model is that we are only able to use data points which contain all necessary parameters for our model. Consequently, we are forced to break our model down into smaller two-dimensional classifiers to maximize the number of training samples.
This obstacle is commonly referred to as the \textit{missing data} problem. Future work will include attempts to address this problem, such as smart imputation via methods like nuclear norm minimization or modified expectation maximization to estimate distributions with missing features.

It is similarly important to note that the normalization method chosen for our model may have a significant impact on its performance. 
As was noted previously, while the logarithmic normalization
makes it easier to handle data points with extremely high or low values
, this may cause the majority of samples to be very close to one another. 
Alternative techniques may be explored to mitigate this issue.

Looking to the future of this work, we aim to identify a data-driven regression strategy to predict
configurations of the actuator (i.e. actuator length, thickness and width, material composition, etc.) that meet the minimum performance metrics. 
We can utilize this regression strategy to 
further separate the nuances between actuator technologies and the selection of actuators. 
This may lead to a unifying strategy for determining the appropriate muscle technology as well as the appropriate configuration for that muscle.

\section*{Acknowledgment}
We would like to thank Jacob John Johnson for helping develop the web application.

\bibliographystyle{IEEEtran}
\balance
\bibliography{bibfile}

\end{document}